\pdfoutput=1

\documentclass[11pt]{article}

\usepackage[]{acl}

\usepackage{times}
\usepackage{latexsym}

\usepackage[T1]{fontenc}

\usepackage[utf8]{inputenc}

\usepackage{graphicx}
\usepackage{caption}
\usepackage{subcaption}
\usepackage{url}
\usepackage{amsmath}
\usepackage{multirow}
\usepackage{colortbl}
\usepackage{float}

\usepackage{comment}
\usepackage{linguex}
\usepackage{booktabs}

\definecolor{darkcerulean}{rgb}{0.03, 0.27, 0.49}
\newcommand{\new}[1]{#1} 

\definecolor{tan}{rgb}{0.69, 0.4, 0.0}

\definecolor{tan}{rgb}{0.69, 0.4, 0.0}
\definecolor{green}{rgb}{0.0, 0.5, 0.0}
\definecolor{bronze}{rgb}{0.8, 0.5, 0.2}
\definecolor{carolinablue}{rgb}{0.6, 0.73, 0.89}
\definecolor{darkgreen}{rgb}{0.12, 0.3, 0.17}
\definecolor{buff}{rgb}{0.94, 0.86, 0.51}
\definecolor{celadon}{rgb}{0.67, 0.88, 0.69}
\definecolor{cherryblossompink}{rgb}{1.0, 0.72, 0.77}
\newcommand{\green}[1]{\textcolor{celadon}{#1}}
\newcommand{\red}[1]{\textcolor{red}{#1}}
\newcommand{\yellow}[1]{\textcolor{buff}{#1}}

\newcommand\Mark[1]{\textsuperscript{#1}}

\usepackage{amssymb}

\usepackage{microtype}

\title{Do Language Models Exhibit Human-like Structural Priming Effects?}

\author{Jaap Jumelet\Mark{1}~~~~~Willem Zuidema\Mark{1}~~~~~Arabella Sinclair\Mark{2} \\[5pt]
\begin{tabular}{c c} 
\Mark{1}Institute for Logic, Language and Computation & \Mark{2}School of Natural and Computing Sciences\\
University of Amsterdam & University of Aberdeen\tabularnewline
\end{tabular}\\
{\small\texttt{jumeletjaap@gmail.com~~~w.h.zuidema@uva.nl~~~arabella.sinclair@abdn.ac.uk}}
}

\begin{document}
\maketitle

\begin{abstract}

We explore which linguistic factors---at the sentence \textit{and} token level---play an important role in influencing language model predictions, and investigate whether these are reflective of results found in 
humans and human corpora \citep{gries_kootstra_2017}. 
We make use of the structural priming paradigm, where recent exposure to a structure facilitates processing of the same structure. We don't only investigate whether, but also \textit{where} priming effects occur, and what factors predict them.
We show that these effects can be explained via the \textit{inverse frequency effect}, known in human priming, where rarer elements within a prime increase priming effects, as well as lexical dependence between prime and target.  
Our results provide an important piece in the puzzle of understanding how properties within their context affect structural prediction in language models.\footnote{All code and data can be found at: \href{https://github.com/jumelet/prime-factors}{github.com/jumelet/prime-factors}.}

\end{abstract}


\section{Introduction}
\label{sec:intro}


Structural priming is the phenomenon where speakers are more likely to repeat a certain structure after being recently exposed to a sentence containing a congruent structure; in the following example a speaker is more likely to produce a \underline{D}ouble \underline{O}bject (\textsc{do}) construction (2a, the \textit{target}) after having been exposed to a sentence with a congruent structure (1a, the \textit{prime}) than after having been exposed to a sentence with an incongruent structure (1b, which illustrates the \underline{P}repositional \underline{O}bject (\textsc{po}) dative construction):

{
\small
\ex.
\a. The girl gave \hspace{2.6 mm}[the boy]$_{NP}$ \hspace{.7 mm}[the ball]$_{NP}$
\b. The girl   gave \hspace{2.6 mm}[the ball]$_{NP}$ \hspace{.7 mm}[to the boy]$_{PP}$

\ex.
\a. The baker gave [the lady]$_{NP}$ [the cake]$_{NP}$

}

Structural priming is well attested in humans, for both language production \citep{mahowald2016meta} and comprehension \citep{tooley2023structural}. Interestingly, it has also been shown to occur in large language models \citep{prasad2019using,sinclair2022structural, michaelov-etal-2023-structural}. 
Here, structural priming can be viewed as a simple form of `in-context learning' \cite{dong2022survey}, where 
the \emph{task} is to generate a sentence (or compute its likelihood) with the target grammatical structure, influenced by the \emph{demonstration} (the \emph{prime} presented to the LLM before processing the target).

%
{Priming effects in humans are typically stronger when there are shared words between prime and target, and when the prime is more unusual, or less frequent.}
This is the \textit{inverse frequency} effect; it extends to other properties of structures themselves, and it is one of the main phenomena we focus on in this paper. 
To explain these effects without direct access to the underlying training data, we turn to factors known to predict priming effects from corpus linguistics~\citep[e.g.][]{gries2005syntactic,JAEGER201357}, which highlight surprisal and structural preference as key factors, and demonstrate the importance of a more fine-grained method of measuring priming.

A second focus of this paper is examining the relationship between lexico-semantic overlap and the asymmetry of the priming effects observed.
We examine priming at the token level, discovering that \textit{where}  priming takes place is important for understanding \textit{how} lexico-semantic factors affect priming and for analysing the mechanisms underlying priming in models.
%
%
Finally, we find that models' structural predictions are highly 
influenced by specific lexical items, and that they incorporate systematic properties of human production preferences learnt from the training data. We demonstrate that models, like humans, exhibit inverse frequency effects in terms of surprisal and verb preference, and that these are predictive of priming.
\section{Structural Priming}
\label{sec:background}
Structural priming in humans is part of a rich literature on factors that impact human language processing, both in controlled experiments of production \citep{mahowald2016meta} and comprehension \citep{tooley2023structural}, and analyses from corpus linguistics ~\cite[e.g.,][]{gries_kootstra_2017}.  
We provide a brief theoretical background on structural priming in \S\ref{sec:priming_background}, and priming in language models in \S\ref{sec:priming_lms}.


\subsection{Properties of Priming}\label{sec:priming_background}

\paragraph{Production vs. Comprehension}
Structural priming has been shown to manifest in both language production and comprehension.
Although recent work has shown that the underlying mechanisms for these two areas may not be as different as originally assumed \citep{SEGAERT2013174,TOOLEY2014101} and are intricately related \citep{dell2014}, numerous works have uncovered distinct differences in the factors that play a role for each modality \citep{ziegler2019use}.
We therefore take the explicit stance in this paper that language models are likely to follow patterns found in human production, since they are exposed solely to human produced data, and for the factors we consider, we find this to be the case.
%
%
%
While LMs are not necessarily expected to align directly with factors found in comprehension studies, 
arguably there may be similar acquisition mechanisms (e.g. error-based learning) that result in comprehension aligned behaviour.
In this background, we focus on production- and corpus-based analyses of structural priming, unless unless explicitly mentioned otherwise.

\paragraph{Inverse Frequency}
One influential theory on the mechanism behind priming in humans is the implicit learning theory by \citet{Chang2006BecomingS}. This theory predicts that our expectation for a particular structure is proportional to degree of \textbf{\textit{surprisal}} of having encountered this structure before.
This effect ---the \textit{inverse frequency effect} (a \textit{rarer} prime will boost priming more) --- has indeed been confirmed experimentally to 
be a strong predictor of 
priming behaviour.
Specifically, in language production in humans
it has been found that highly surprising primes (as measured by language models) will have higher priming effects~ \citep{grieswulff2005,Jaeger2008cog,JAEGER201357,languages9040147}. 

Relatedly, \textbf{\textit{structural preference}}---which expresses within which structure 
a verb is most likely to occur---is another important factor when predicting priming behaviour: 
%
verbs that are strongly associated with one construction are more likely to be primed by that construction as well \citep{grieswulff2005, GriesHampeSchönefeld+2005+635+676, BERNOLET2010455}. From this it then follows that priming effects are stronger when the prime sentence was of a less preferred structure: a prime containing the verb \textit{gave}, for example, will prime subsequent targets more strongly when it is encountered in its \textit{dispreferred} structure (\textsc{po}) \citep{pickering1998representation,zhou2023affects}.
There exists an extensive line of work into determining the factors that govern this structural preference, which is driven by various complex syntax-semantic interactions \citep{Green1974SemanticsAS, THOMPSON1987399, gropen1989, bresnan2007predicting}.
Inspired by this literature, we find evidence in \S\ref{sec:factors} of a verb-mediated inverse frequency effect in modern LLMs.



\paragraph{Lexical Dependence}
Many findings in production and corpus studies have shown that priming effects of sounds, words, meanings and structures interact:
prime sentences and target sentences with shared words (\textbf{\textit{lexical overlap}}), or words that share semantics (\textbf{\textit{semantic overlap}}), boost structural priming \citep{hare1999structural,jones2006high, HARTSUIKER2008214, snider2009, gerard2010corpus}, and similar findings have been found in comprehension studies as well \citep{chiarello1990semantic, lucas2000semantic, traxler2014syntactic}.
A common explanation is that words in the prime that are identical or similar to words in the target already activate the relevant abstract syntactic frames. These frames, in turn, are most closely associated with verbs, or the syntactic head of the primed structure~\citep{pickering1998representation, pickering2008structural, reitter2011computational}. 

Lexical overlap effects in human experiments typically do not consider effect of preposition or determiner overlap, rather focusing on the content words. Findings have shown that structural priming does not depend on the repetition of function words, thus in humans there is a clear difference between content-word and function word repetition~\citep{bock1989closed,tree1999building,pickering2008structural}. 

\subsection{Structural Priming in Language Models}\label{sec:priming_lms}
Structural priming has been used to investigate abstract language representations in language models. A number of (early) papers used fine-tuning on a small sample of items of a particular structure, and measured its impact on related items \citep{van-schijndel-linzen-2018-neural,prasad2019using}.
%
\citet{sinclair2022structural} measure the impact of congruent and incongruent prime sentences on a subsequent target, paralleling approaches in psycholinguistics that view priming as resulting from \textit{residual activations} \citep{branigan1999syntactic}. 
Using this approach, LMs are shown to exhibit priming effects  that are cumulative, susceptible to recency effects, boosted by lexico-semantic overlap, and persisting in cross-lingual settings \citep{michaelov-etal-2023-structural, xiao2024modeling}. 


One key finding of \citet{sinclair2022structural} is that priming effects are often asymmetric: when comparing alternative structures in the dative and transitive data, they remark that some of these structures are more susceptible to priming than their alternatives. In \S\ref{sec:experiments} we confirm this observation for the dative; the strength and direction of the asymmetry are a surprising result, given priming effects are typically higher for the \textit{opposite} alternation in humans~\citep{bock1989closed,kaschak2011structural,reitter2011computational}. We show that this finding extends to a wide range of state-of-the-art LLMs, and is predictable via other inverse frequency effects.

\section{Measures, Data \& Models} 

\subsection{Sentence-level Priming Effect}

To measure the priming effect, we make use of the measure of \citet{sinclair2022structural}, which has recently also been adapted by \citet{sinha-etal-2023-language} and \citet{michaelov-etal-2023-structural}. 
The {Priming Effect} (PE) is defined as the difference in log probability of a target sentence $\textsc{t}^\textsc{x}$ when preceded by a prime $\textsc{p}^\textsc{x}$ that has the same congruent structure $\textsc{x}$ (\textsc{po}/\textsc{do}), and the log probability of the same target $\textsc{t}^\textsc{x}$ that is preceded by a prime $\textsc{p}^\textsc{y}$ of incongruent structure $\textsc{y}$ (to contrast this measure with the measure from \S\ref{sec:token-pe}, we will refer to it as the \emph{sentence}-level Priming Effect, $s$-PE): 
\begin{equation}\label{eq:pe}
    s\textup{-}PE(\textsc{x}) = \log P(\textsc{t}^\textsc{x}|\textsc{p}^\textsc{x}) - \log P(\textsc{t}^\textsc{x}|\textsc{p}^\textsc{y})
\end{equation}
The conditional probability of $\log P(\textsc{t}^\textsc{x}|\textsc{p}^\textsc{x})$ is computed as the sum of log probabilities of all tokens in the target sentence:
\begin{equation}
    \log P(\textsc{t}^\textsc{x}|\textsc{p}^\textsc{x}) = \sum_i \log P_{LM}(\textsc{t}^\textsc{x}_i|\textsc{p}^\textsc{x},\textsc{t}^{\textsc{x}}_{<i})
\end{equation}


\subsection{Token-level Priming Effect}\label{sec:token-pe}
The Priming Effect metric of Eq. \ref{eq:pe} shows whether a target sentence is primed by structural congruence as a whole, but does not provide insight into \textit{which} tokens within the target were most responsible for such an effect.
To investigate this, we introduce the \textbf{token-level priming effect} metric ($w\textup{-}PE$), which expresses priming effects for each individual target token $T_i^\textsc{x}$:
\begin{align}\label{eq:token-pe}
\begin{split}
    w\textup{-}PE(\textsc{x}, i) &= \log P(\textsc{t}^\textsc{x}_i|\textsc{p}^\textsc{x}, \textsc{t}^\textsc{x}_{<i}) \\&- \log P(\textsc{t}^\textsc{x}_i|\textsc{p}^\textsc{y}, \textsc{t}^\textsc{x}_{<i})
\end{split}
\end{align}
Note that the sentence-level PE decomposes into a sum of $w$-PE scores; as such $w$-PE expresses the relative contribution of each target token to $s$-PE:
\[s\textup{-}PE(\textsc{x}) = \sum_iw\textup{-}PE(\textsc{x}, i)\]


\subsection{The Prime-LM Corpus}
\label{sec:corpus}
We use the dative constructions from the Prime-LM corpus of \citet{sinclair2022structural}, similar to examples (1) and (2) in \S\ref{sec:intro}. This subset of sentences is convenient for our purposes, because we can select both prime-target pairs with
\textit{no} lexical overlap and minimal semantic similarity between nouns and verbs, as well as pairs with varying degrees of overlap and varying degrees of semantic similarity. The datives thus allow us to not only measure structural priming, but also inspect the role of lexical overlap in more detail. 
We briefly explain the subsets we select for our experiments (each containing 15.000 prime/target pairs), as well as two additional sub-conditions we introduce to the lexical overlap category.

\begin{description}
\item[Core] 
contains
a) no lexical overlap exists between prime and target sentences, not even between function words, and b) 
no semantic association exists between prime and target exists in the USF free association norms dataset \citep{nelson2004university}.
In our experiments, we use the Core condition as a baseline. 
    
\item[Semantic Similarity] 
contains explicit pairwise semantic similarity between prime and target,  
where similarity is assessed by a non-zero human association from the USF dataset or a minimum cosine similarity of at least 0.4 based on GPT2-large embeddings.
We consider three conditions
: i) all nouns are semantically similar, ii) the verbs are similar, iii) all nouns \textit{and} verbs are similar.

\item[Lexical Overlap] 
ensures lexical items are shared across prime and target.
We consider three such conditions
: i) all nouns overlap, ii) determiners and prepositions overlap, iii) verbs overlap. We create two additional conditions, iv) determiner overlap and v) preposition overlap. 
This allows us to separately measure the impact of determiners and prepositions, 
since \textit{verb overlap} necessitates preposition overlap. 
\end{description}

\begin{figure*}[hbt!]
    \centering
    \includegraphics[height=1.65in]{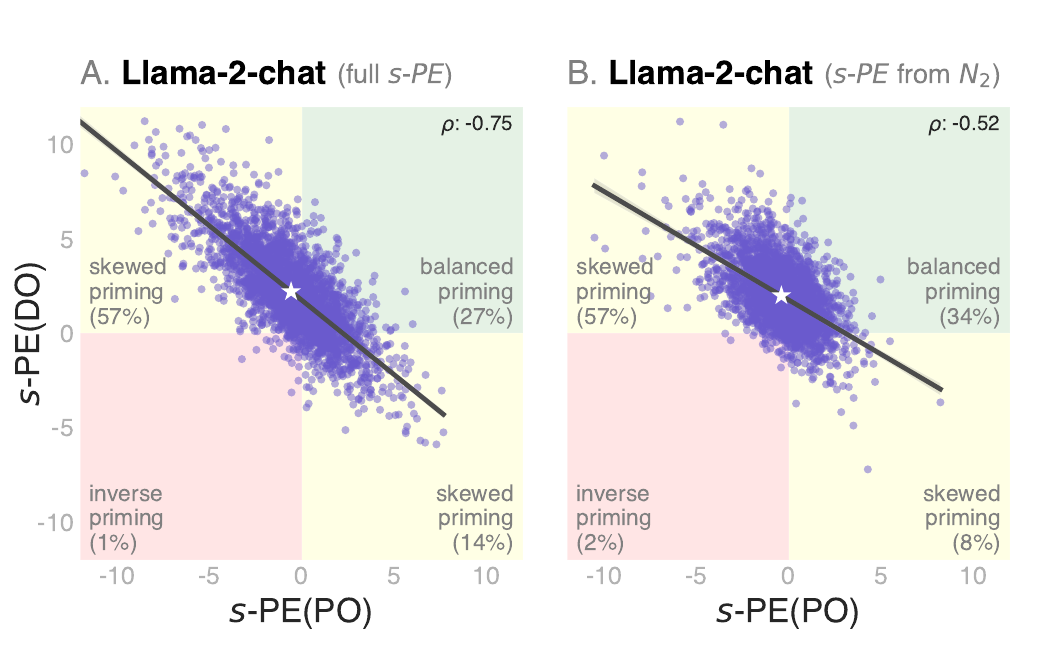}\\
    \includegraphics[height=1.65in]{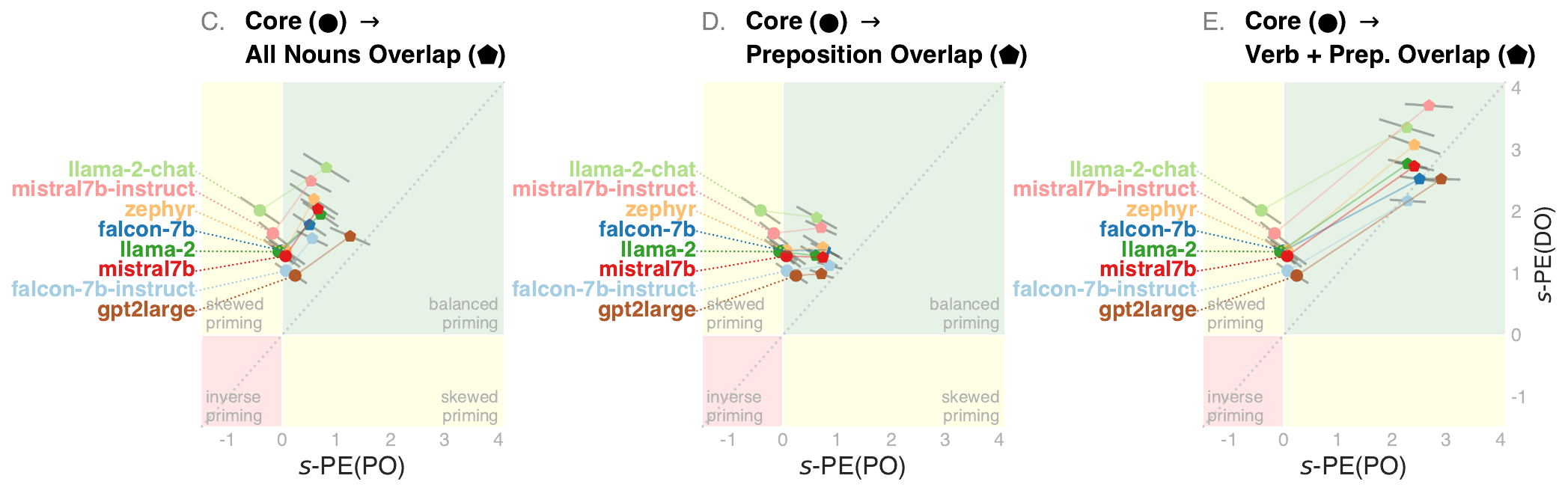}
    \caption{
    We plot PE results against one another. The four quadrants in this `\textit{PE space}': \textit{\green{balanced} priming} where the PE is positive in both directions, \textit{\yellow{skewed} priming} where it's only positive in one, and \textit{\red{inverse} priming} when the PE is negative in both directions.
    There exists a strong negative correlation between priming effects of opposite structures (A). 
    Only a small portion of the data is primed in both directions for Core. 
    Priming becomes more balanced when measured from the point of divergence in the target (B, \S\ref{sec:token-pe-experiments}), or when lexical overlap is increased (C--E).
    }
    \label{fig:sentence_level}
\end{figure*}

\subsection{Models}\label{sec:models} 
We consider the following (auto-regressive) LLMs. 
For models with an \textsuperscript{*} we also test their \textit{aligned} versions.
PE scores are computed using the \texttt{diagnnose} library \citep{jumelet-2020-diagnnose}.
\begin{description}
    \item \textbf{GPT2-large} \citep{radfordlanguage}: This is the impactful 2019 model from OpenAI, with 774M parameters, trained only on a (causal) language modelling objective.
    \item $^*$\textbf{Llama-2-7b} \citep{DBLP:journals/corr/abs-2307-09288}: We consider both the 7B base model and the RLHF/PPO aligned \textit{chat} model \citep{DBLP:conf/nips/Ouyang0JAWMZASR22}.
    \item $^*$\textbf{Falcon-7b} \citep{DBLP:journals/corr/abs-2311-16867}: We consider both the 7B base, and the \textit{instruction-tuned} variant fine-tuned on dialogue data taken from ChatGPT \citep{DBLP:journals/corr/abs-2303-08774}.
    \item $^*$\textbf{Mistral-7b} \citep{DBLP:journals/corr/abs-2310-06825}: This 7B model is the current state-of-the-art in this size bracket. We also consider the instruction-tuned variant, trained on similar data to Falcon-7b.
    \item $^*$\textbf{Zephyr} \citep{DBLP:journals/corr/abs-2310-16944}: An aligned version of Mistral-7b 
    using Direct Preference Optimization \citep{DBLP:journals/corr/abs-2305-18290}.
\end{description}

\section{Exp1: Measuring Structural Priming}\label{sec:experiments}
We aim to better understand the asymmetrical priming effects observed in \citet{sinclair2022structural}, to gain a more detailed picture of how lexical overlap affects this asymmetry.
We start our experimental setup with their sentence-level approach, considering a wider range of large, contemporary LLMs. 
In the next section we then examine priming effects at a more fine-grained level. 




\paragraph{{Priming Effects are skewed and correlated}}
We compute the $s$-PE scores for the models of \S\ref{sec:models} on the Core condition of Prime-LM.
We observe there exists a strong negative correlation between the PE scores of the prepositional object and the double object.
In Figure~\ref{fig:sentence_level}A we plot those results 
as a scatter plot in the space formed by PE score for one construction against PE score for the alternative construction. This representation highlights that, 
%
for Llama-2-Chat and all the other models we consider, the $s$-PE of \textsc{po} constructions is \emph{negatively} correlated with that of \textsc{do} constructions ($\rho$: -0.72 to -0.77), and that only for a fraction of sentences there exists a positive priming effect in both directions (26 to 38\%).

{This correlation and skew towards one of the two constructions were already observed by \citet{sinclair2022structural} for GPT2-large and other relatively small LMs. Interestingly, correlation and skew do also exist in the newest, large LMs, and, moreover, are even more pronounced. Figure~\ref{fig:sentence_level}C shows the mean $s$-PE in the same \textit{PE space} for all the LLMs we considered. 
GPT2-large shows the least skew, Llama-2-chat the most (for completeness, the plot also shows the strength of the correlation between the PE(\textsc{po}) and PE(\textsc{do}) scores, as well as the spread of the distribution.)} 
Note that this observed behaviour of \emph{large} LMs is less consistent and far more asymmetric than results in the human literature, where priming effects, while typically asymmetric~\citep{bock1989closed} 
, are generally observed to be positive for both structures.



\paragraph{{Lexical overlap balances Priming Effects}}

Next, we investigate the priming effects of the LMs where either the semantic similarity or lexical overlap between prime and target is increased.
The results for lexical overlap are shown in Figure~\ref{fig:sentence_level}C--E (additional plots regarding semantic similarity are in Appendix~\ref{app:semsim_spe}). 
The plots show that an increase in lexical overlap of any type moves all models more solidly into the upper-right quadrant of the PE-space. 
That is, it pushes all LM priming behaviours to become both stronger and more balanced {(less skewed towards one or the other construction)}. 
This is especially prevalent for the overlap in verbs and function words. 
We will explore the impact of these factors in more detail in the next section.

\section{Exp2: Locating Structural Priming} 
\label{sec:token-pe-experiments}

\begin{figure*}[ht!]
    \centering
    \includegraphics[width=\textwidth, clip, trim={0 0 6cm 0}]{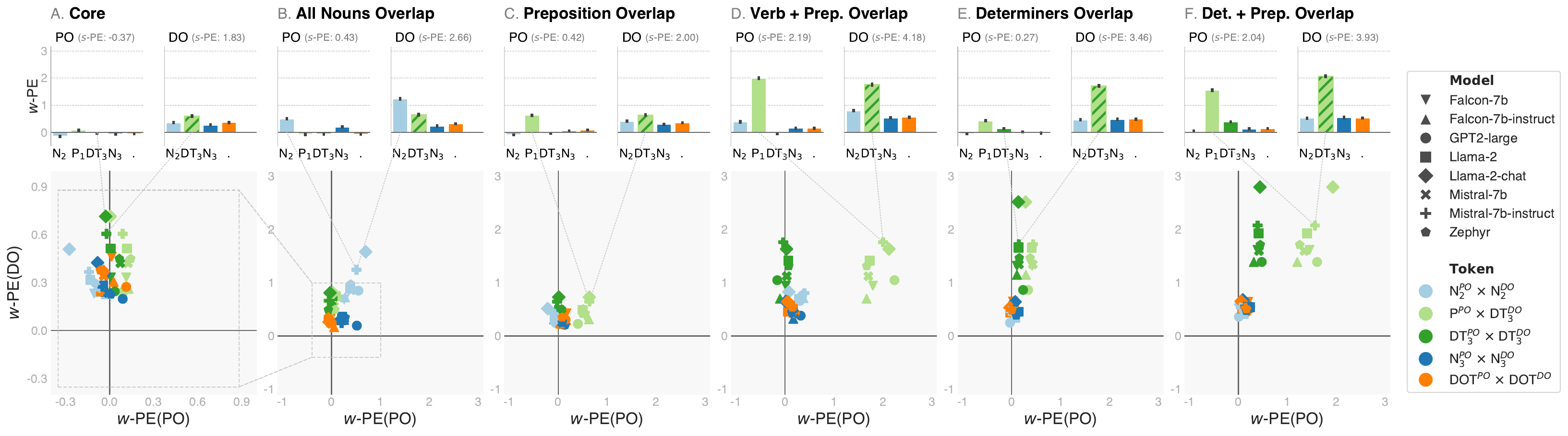}\\
    \includegraphics[width=\textwidth,clip,trim={1cm 1.2cm 0 0}]{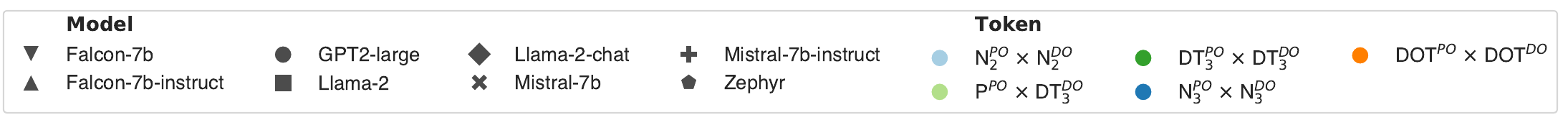}
    \caption{
    The $w$-PE scores for the \textit{Core} and \textit{Lexical Overlap} conditions.
    Scores are grouped by token (based on colour) and model (based on shape).
    To exemplify how these \textit{Priming Space} coordinates map to a bar chart, we show the Mistral-7b-instruct scores at the top of each plot.
    Note that the \textit{Core} results are plotted at a different scale than the other conditions.
    \textsc{po}: \textcolor{lightgray}{\textit{The girl gave the}} \textcolor[HTML]{A6CEE3}{\textbf{ball}} \textcolor[HTML]{B2DF8A}{\underline{to}} \textcolor{green}{the} \textcolor[HTML]{1F78B4}{\textbf{boy}} \textcolor{orange}{.} 
    \textsc{do}: \textcolor{lightgray}{\textit{The girl gave the}} \textcolor[HTML]{A6CEE3}{\textbf{boy}} \textcolor[HTML]{33A02C}{\underline{the}} \textcolor[HTML]{1F78B4}{\textbf{ball}} \textcolor{orange}{.} 
    }
    \label{fig:full_token_pe}
\end{figure*}

\begin{figure}
    \centering
    \includegraphics[width=\columnwidth,trim={0cm 0 1.0cm 0},clip]{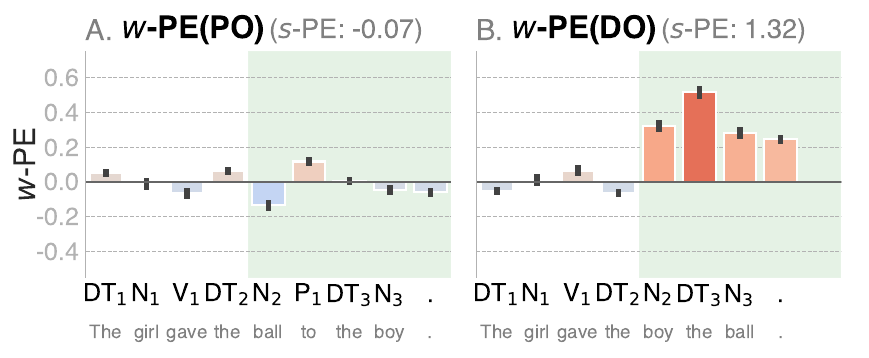}
    \caption{
    The token-level Priming Effect reveals which token predictions in the target sentence contributed the most to the overall sentence-level Priming Effect, here averaged for Llama-2 over the \textit{Core}.
    It is inversely correlated up to the point of divergence between the two structures, at the position of the second noun (\textsc{n}$_2$).
    Only after that point can the congruence between prime and target start playing a role.
    }
    \label{fig:wpe-example}
\end{figure}

Sentence-level analysis does not allow us to investigate individual token-level predictions, making it impossible to examine \textit{where} in the target sentence priming effects are at their strongest.
To better understand the $s$-PE results of \S\ref{sec:experiments} we thus compute the token-level $w$-PE scores for the same conditions.

\paragraph{Structural Divergence} 
Figure~\ref{fig:wpe-example} shows the average $w$-PE scores for Llama-2 on the \textit{Core} condition, which exhibits much higher sentence-level priming effects for the \textsc{do} sentence than for the \textsc{po} sentence. The \emph{token-level} scores show that the treatment of the two sentences starts to diverge
from the position of the \textit{second} noun {onwards (the second noun is is the patient for \textsc{po}, whereas it is the recipient for \textsc{do}).} Prior to that, the target sentence is, in fact, the same for both \textsc{po}/\textsc{do} alternations and as such will be inversely proportional to each other: scores up till this point merely show a target's bias towards a prime of a particular structure, regardless of structural congruence.

This provides a partial explanation for the strong negative correlation between $s$-PE scores that was observed in \S\ref{sec:experiments}: since (roughly) half of $s$-PE score is made up of $w$-PE scores that have a perfectly negative correlation of -1, the overall correlation of $s$-PE scores is strongly affected by this. 
Based on this insight we compute a modified $s$-PE score that is only measured from the point of target sentence divergence ($s_\delta$):
\begin{equation}\label{eq:sdelta_pe}
    s_\delta\textup{-}PE(\textsc{x}) = \sum_{i>4}w\textup{-}PE(\textsc{x}, i)
\end{equation}
This allows us to confirm that the negative correlation between \textsc{po} and \textsc{do} decreases with $s_\delta\textup{-}PE$ ($\rho$: -0.76 to -0.52 for Llama-2-chat; Figure~\ref{fig:sentence_level}B).

\paragraph{Lexical Dependence} 
%
%
We focus our analysis on \textit{lexical overlap} (Figure~\ref{fig:full_token_pe}), which showed the strongest balancing effects in \S\ref{sec:experiments}.\footnote{
    Results for semantic similarity are provided in Appendix~\ref{app:semsim_wpe}.
} 
Priming behaviour could be distributed in two ways across the target: either uniformly or peaked at a particular token, and either balanced or skewed towards one structure.
From the point of structural divergence, we have the noun (N$_2$) of the first noun phrase, followed by the function word (P$_1$:\textsc{po}, DT$_3$:\textsc{do}) that marks the start of the second noun phrase of the construction. 
Priming effects for N$_2$ stem from cues with respect to the \textit{semantic role} (e.g. \textcolor{lightgray}{gave the} \textcolor[HTML]{1F78B4}{ball}$_{\textsc{po}}$ $ |$ \textcolor[HTML]{1F78B4}{boy}$_{\textsc{do}}$). 
Numerous works in production have shown priming to already take place at this location \citep{pickering1998representation,cleland2003use}.
We would therefore expect to find some evidence of \textit{balanced} priming from the N$_2$ within the core condition. 
However, the \textit{skewed} priming we observe in \textit{Core} (\ref{fig:full_token_pe}A) suggests that the semantic role of the noun does not play as important a part in structural prediction for models. 
Indeed, we observe the most consistent and balanced priming effects from the start of the second NP ($w$-PE(\textsc{po}, \textsc{p}$_1$), $w$-PE(\textsc{do}, \textsc{dt}$_3$)), suggesting that models only narrow their structural predictions later on within a sentence.

Next, we observe the local impact of lexical overlap between prime and target.
For overlapping nouns, we can see that the $w$-PE for both \textsc{n}$_2$ and \textsc{n}$_3$ has increased significantly for both \textsc{po} and \textsc{do}.
The other tokens, on the other hand, are not impacted by this overlap at all: the priming boost manifests itself solely at the position of the overlapping token.
%
For verb overlap, we show that the increased $s-$PE scores here stem from the verb as well as the prepositional overlap (necessary when sharing the same verb and preserving semantics) resulting in  significantly larger $w$-PE(\textsc{po}, \textsc{p}) scores (Figures~\ref{fig:full_token_pe}C and D). 
Interestingly, verb overlap also leads to a boost in \textsc{n}$_2$ and \textsc{n}$_3$, compared to the \textit{Core}.
This shows that, under this condition, the model \textit{is} aware of the expected semantic role in the \textsc{n}$_2$ position: the verb overlap has primed the model in the \textsc{do} case to expect an animate entity here (and inanimate for \textsc{po}).
Unlike findings in the human literature for both production~\citep{bock1989closed} and comprehension ~\citep{traxler2008structural}, we observe prepositional overlap strongly boosting priming effects in the language models we investigate.

\section{Exp3: Explaining Structural Priming}\label{sec:factors}

We now take a closer look at the factors that impact Priming Effects by conducting a regression analysis inspired by factors from corpus linguistics and production studies \citep{gries2005syntactic,JAEGER201357}.
Following \citet{gries2011}, we make use of linear mixed effects models to determine salient word and sentence level factors that predict priming, 
%
%
to discover whether models display consistent behaviour with respect to one another and to human patterns of priming in production they may have learnt.\
We first describe the factors we use in our regression analysis in \S\ref{sec:priming_factors}, and present the results in \S\ref{sec:lmm_results}.

\subsection{Priming Factors}\label{sec:priming_factors}

We investigate the two broad categories of factors discussed in~\S\ref{sec:background}: \textit{lexical dependence}, making use of the various conditions of the Prime-LM corpus (\S\ref{sec:corpus}), and \textit{inverse frequency}, choosing to focus on sentence-level surprisal and the structural preference of the verbs used. 

\paragraph{Lexical Dependence} We include pairwise token-level \textbf{\textit{semantic similarity}} across prime and target content words, measured as the cosine similarity of the word embeddings taken from GPT2-large.
We also include sentence-level similarity, based on the sentence embeddings of MPNet \citep{DBLP:conf/nips/Song0QLL20}, a high-performance sentence encoder. Here, we compute the cosine similarity between the \textsc{po} prime and target embeddings.
We add \textit{\textbf{lexical overlap}} {as a binary factor per token to our analysis.} This allows us to separate overlap effects in conditions where multiple tokens overlap, which is not possible in corpus-level experiments.

\paragraph{Surprisal}
We include the \textit{\textbf{surprisal}} of the congruent and incongruent prime and target, based on the negative log likelihood of the language model itself. 
Surprisal gives us a measure of how predictable or expected the sentences are as a whole, encompassing within-sentence collocation frequency effects.

\paragraph{Structural Preference}\label{sec:struct_pref}


Whereas corpus-based analysis of preferences is often based on normalised frequency statistics \citep{gries2004collexeme}, we base preference on the average probability difference of a verb in two alternating structures:
\begin{equation}
    \textsc{po}\textit{-pref}(v) = \frac{1}{|\mathcal{V}|}\sum_{s\in \mathcal{V}}\log P(s^{\textsc{po}}) - \log P(s^{\textsc{do}})
\end{equation}
where $\mathcal{V}$ is the set of sentences containing verb $v$. 
This score expresses a verb's preference towards a particular structure on a scale from \textsc{do} to \textsc{po}.
\citet{hawkins-etal-2020-investigating} and \citet{veenboer-bloem-2023-using} provide a similar methodology for measuring structural preferences in LMs.
For computing these scores we make use of the prime sentences from the \textit{Core} condition of PrimeLM.

\begin{figure}[h!]
    \centering
    \includegraphics[width=\columnwidth,trim={0.2cm 0 0.3cm 0},clip]{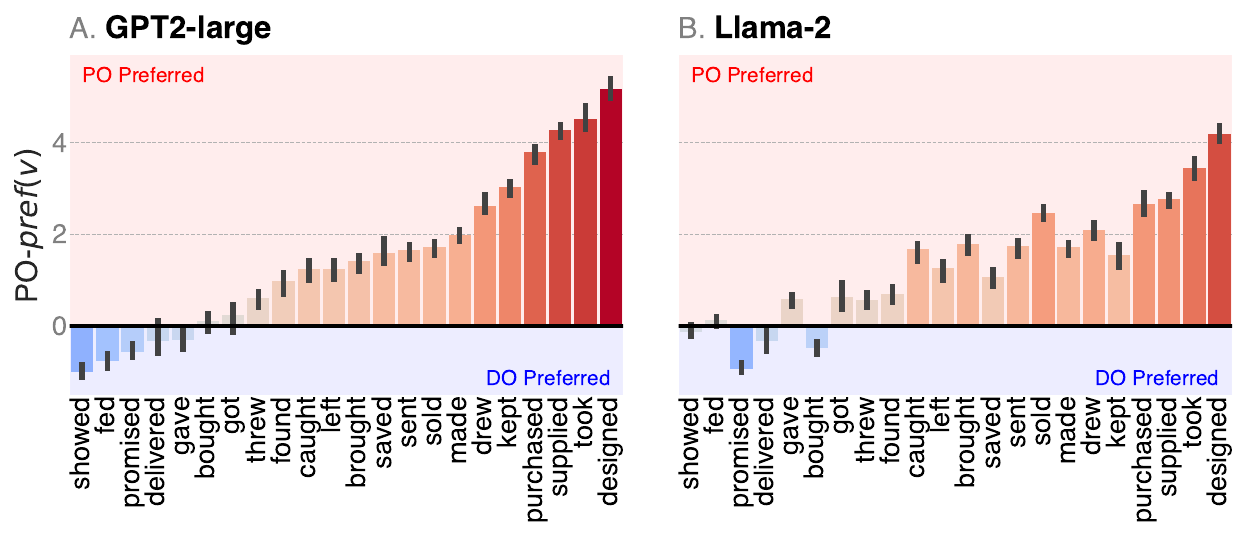}    
    \caption{
    Structural preferences for GPT2-large and Llama-2, expressing the preference of a ditransitive verb with respect to a prepositional object versus a double object construction.
    The verb order of Llama-2 is based on the sorted order of GPT2-large.
    }
    \label{fig:structural_preference}
\end{figure}
%
The majority of verbs have a preference towards \textsc{po} structure (as an example, Figure \ref{fig:structural_preference} contains the preferences of GPT2-large and Llama-2).
This is not in line with dative {usage} preferences found in English, although it varies across vernaculars: {some production and corpus studies  
suggest American English has a 2:1 preference towards DO constructions \citet{bock2000persistence,grimm2009spatiotemporal}, whereas Australian English has a PO preference \citep{bresnan2010predictingsyntax}.}
Preference towards \textsc{po} in LMs may be confounded by transitive verb phrases followed by a prepositional modifier.


We also compute the Spearman correlations between the preference orders of all LMs and humans \citep{gries2004collexeme}, which reveals that there exists a high degree of variance across models and low correlation across models and human preference order (full figure in Appendix~\ref{sec:structural_preference_correlations}).
We leave a more thorough investigation of these differences open for future work that can take inspiration from established linguistic findings \citep{gropen1989, heaviness2000}.


\subsection{Modeling Priming Effects}\label{sec:lmm_results}
\paragraph{Linear Mixed Model}
We fit a linear mixed model (LMM) using the factors of \S\ref{sec:priming_factors} that are added as fixed effects \citep{BAAYEN2008390}.
We fit two LMMs: one for predicting $s_\delta$-PE(\textsc{po}) and one for $s_\delta$-PE(\textsc{do}), which will provide insights whether different factors predict these effects.
Fitting is done based on 30.000 items: 15.000 items are sampled for the \textit{Core} condition, and 15.000 items are sample from the \textit{Semantic Similarity} and \textit{Lexical Overlap} conditions.
This provides a balanced dataset of the \textit{Core} and conditions that diverge from this baseline.
We add a by-LM random intercept to account for individual model biases, akin to by-speaker random effects in human priming studies \citep{gries2011, JAEGER201357}.
All factors are centred and scaled to unit variance. 

\paragraph{Results}
\begin{figure}
    \centering
    \includegraphics[width=0.9\columnwidth]{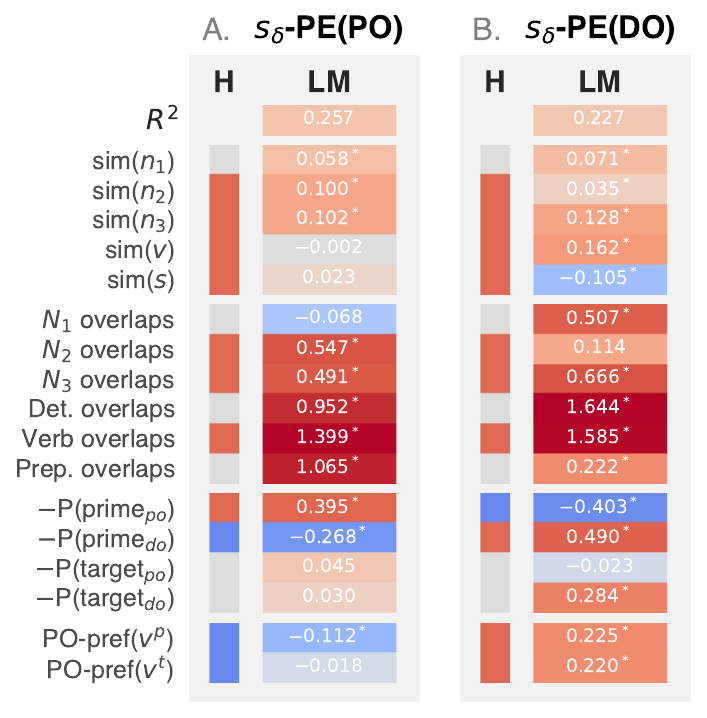}
    \caption{
    LMM coefficients for (A) predicting $s_\delta$-PE(\textsc{po}) and (B) $s_\delta$-PE(\textsc{do}), shown side-by-side with reported effects for predicting human priming in production- and corpus-based studies.
    Significant LLM coefficients ($p<10^{-3}$) are denoted by an asterisk.
    }
    \label{fig:pe_factors}
\end{figure}
We report full LMM with coefficients in Figure~\ref{fig:pe_factors}, next to the reported effects found in the literature on human priming ($z$-scores and standard errors in Appendix~\ref{app:lmm_summary}).
The LMM reaches an $R^2$ of 0.257 (\textsc{po}) and 0.227 (\textsc{do}), which indicates that a large fraction of PE could still be predicted based on other factors and more complex interactions.
We leave a more extensive exploratory analysis for future work, and focus on confirmatory hypothesis testing for now \citep{tukey1980, BARR2013255}.

\paragraph{Semantic Similarity}
As in human production and corpus linguistics findings, we observe priming effects are 
predicted by semantic similarity between prime and target words~\citep{snider2009}, \new{with the most consistent effects across structures for noun similarity~\citep{cleland2003use}}, although the effect is relatively weak compared to other factors. 
Sentence similarity however, was not a predictive factor.
In part this could be due to the sentence encoder not being sensitive to the structurally similar \textsc{po} items that we use for computing sentence-level similarity.
Incorporating the feature-based Grower similarity employed by \citet{snider2009} could be an alternative to explore the relation between sentence-level semantic similarity and priming.

\paragraph{Lexical Overlap}
We observe that lexical overlap is the strongest predictor of priming behaviour, in particular for primes sharing the same verbs, prepositions and determiners as the targets. 
This is in line with human findings; the meta analysis of \citet{mahowald2016meta} shows lexical overlap is `the most consistent moderator of syntactic priming'.
We  also observe that shared determiner overlap is consistently of high importance when predicting model PE, something observed but given far less attention in the human literature.
Contrary to human findings in both production~\citep[e.g.][]{bock1989closed} and comprehension~\citep[e.g.][]{traxler2008structural}, prepositional overlap is one of the strongest priming predictors.
This indicates that priming in LMs is strongly driven by lexical cues, tying in with our observation in \S\ref{sec:token-pe-experiments} that priming effects are highly influened by single token prediction, and this is 
driven more strongly by function than content words.

\paragraph{Surprisal}
Similar to \citet{JAEGER201357} and \citet{languages9040147}, we find that priming effect is predicted by prime surprisal, in both directions for \textsc{po} and \textsc{do}.
This is evidence for an {inverse frequency effect}: a less frequent/plausible prime leads to an \textit{increase} in priming effect. 
Target surprisal is less significant: only \textsc{do} surprisal is a significant predictor. 


\paragraph{Structural Preference}
We find that verb preference plays a highly predictive role, which again provides evidence for inverse frequency effects.
A verb that has a structural preference for \textsc{po} will lead to a higher \textsc{do} priming effect, and vice versa, in line with results observed in human data~\citep{grieswulff2005, GriesHampeSchönefeld+2005+635+676}.
This provides further explanation for the \textsc{do} skewed priming effects that models display: for most of \textsc{do} targets, their primes will \textit{not} be in the preferred structure, thus boosting priming effects. 
\section{Discussion \& Conclusion}
In this paper, we seek to better understand the mechanisms that may underlie structural priming behaviour in LLMs. 
Borrowing insights from empirical and theoretical work on priming in humans, we investigate how, where and why a range of modern LLMs assign higher or lower probabilities to target sentences depending on preceding context, allowing us to investigate the extent to which language models are influenced by structure and semantics when making upcoming predictions. 


\paragraph{Do models demonstrate structural priming?}
We find, in line with~\citet{sinclair2022structural}, that models exhibit asymmetrical priming effects, and that this is even more pronounced in newer, larger LMs. 
By introducing a token-level priming effect we are able to locate more precisely potential sources of this asymmetry.
We observe the direction of the asymmetry in PE is consistently \textit{inverse} to priming effects in humans: where humans consistently display higher PE for the \textsc{po} alternation, rather than \textsc{do}, which we observe in models. 
%
We speculate that the verb preference effects we find in \S\ref{sec:factors}, which are predictive of PE as in humans, may play a role in this. 
Finally, through observing priming at the token level, we observe that balanced priming in models is only visible from later on within the target, later than we may expect the same effects to be observed in humans.


\paragraph{How do humans and models compare?}
In predicting model PE using human factors known to predict priming in humans and in human corpora, we observe that lexico-semantic and frequency-related predictors of priming in humans also predict priming in LLMs. 
However, although we find lexico-semantic overlap of content words to be a reliable predictor of priming, we find that \textit{function word overlap} plays a surprisingly predictive role, which has not been shown to be the case for humans with dative constructions~\citep{bock1989closed,pickering2008structural}.
Similar to human findings, we observe---consistently across models---inverse frequency effects of prime surprisal and verb preference~\citep[e.g.][]{grieswulff2005,Jaeger2008cog}. 
This demonstrates that models are able to pick up on highly abstract factors influencing language predictions in humans from corpora, and highlights how influential seemingly small properties of the context are when it comes to upcoming model predictions.

\paragraph{What are the implications of implicit learning?}
We showed that priming is driven by similar inverse frequency effects observed in human priming.
From a broader perspective, this is a striking finding.
Inverse frequency effects have been argued to stem from an error-based implicit learning procedure \citep{Chang2006BecomingS}: we adapt future predictions proportionally to recent predictive errors.
This cognitive mechanism then leads to detectable patterns in human-produced corpus data \citep{JAEGER201357}, on which LLMs are trained.
LLMs are thus able to pick up on this highly abstract pattern, which shows that their priming behaviour is far more intricate than a simple repetition-based mechanism.
An interesting endeavour for future work would be to test this finding in a setting with control over data distribution (e.g. \citet{jumelet-zuidema-2023-transparency}), to ensure that inverse frequency effects do not stem from some other indirect effect of language learning.

\paragraph{Comprehension and production in LLMs}
We build on \citet{sinclair2022structural}, who design the priming effect metric to measure \textit{comprehension} in LLMs forcing its prediction on a fixed target without allowing for open-ended production.
It is important to remember, however, that through the way LLMs are trained, their predictions will be driven by human \textit{production} patterns in the training data, thus motivating our choice to base our predictions on findings from corpus and production studies.
Although the mechanisms for comprehension and production in LLMs are highly linked---they rely on the same output distribution---it would be interesting to investigate priming in LLMs in a generation-based setting as well.
A thorough investigation in this direction may provide deeper insights into the relation between LLM behaviour and cognitive theories of human language processing \citep[e.g.,][]{dell2014}.

\paragraph{Outlook}
Language models as cognitive models can potentially aid in discovering important properties of human linguistic behaviour~\cite{futrell-etal-2019-neural,linzen2021syntactic,10.1162/ling_a_00491}; we thus view our results as a contribution to defining the border where human patterns are replicated in models.
Looking outwards from this detailed analysis, future studies could investigate the extent to which these priming effects influence structural repetition patterns in generation, complementing existing work finding priming-like lexical repetition effects in LLM generation~\cite{molnar2023attribution}.
Furthermore, a more detailed investigation in the exact nature of the (potentially) hierarchical representations that underlie priming behaviour could take inspiration from parsing-based theories of priming \citep{prasad2024spawning}, or deploy techniques from interpretability research to uncover hierarchical structure \citep{DBLP:conf/iclr/MurtySAM23, jumelet-zuidema-2023-feature}.
Not only will such investigations provide deeper insights into the cognitive plausibility of LLMs \citep{beinborn2023cognitive}, but it may also yield a better understanding of the mechanisms underlying in-context learning \citep{min-etal-2022-rethinking, han-etal-2023-understanding}.

\section*{Limitations}
One clear limitation of our work stems from the specific nature of our analyses. More work remains to be done to investigate whether these results generalise across other constructions within English or further extend to other languages. It also remains an open question whether the constraints within the dataset we use influence our outcomes: do these results generalise to a wider range of vocabulary, or a more complex set of sentences.
Additionally, even within the scope of our analyses, the effects of model size on the results we observe are interesting, but would need to be tested more systematically to draw firm conclusions from them. Likewise, while we purposefully include a selection of base models and their alignment tuned variants to investigate whether there are any striking differences, the sample is too small to make any meaningful inference.

\section*{Acknowledgements}
We would like to thank our three anonymous reviewers and meta-reviewer for their thoughtful comments and discussion, which helped refine this paper. 
We would like to thank Raquel Fernández for her fundamental role in the paper upon which this paper builds.
AS would like to thank colleagues Anastasia Klimovich-Gray and Agnieszka Konopka from the department of Psychology for their useful discussions and psycholinguistic perspective.  
JJ would like to thank Jakub Szymanik, Roberto Zamparelli, and Alexey Koshevoy for insightful discussions during his Trento visit.


\bibliography{anthology,struct_prime}
\bibliographystyle{acl_natbib}

\appendix

\section{Author Contributions}
Following the Contributor Role Taxonomy (CRediT; \citealt{allen2019}):

{\small \centering
\begin{tabular}{r|l l l}
     & JJ & WZ & AS \\\hline
    Conceptualisation & \checkmark & \checkmark & \checkmark\\
    Methodology & \checkmark &  & \checkmark\\
    Software & \checkmark &  & \\
    Data Curation & \checkmark & & \checkmark \\
    Investigation & \checkmark &  & \\
    Visualisation & \checkmark &  & \\
    Analysis & \checkmark &  & \checkmark\\
    Writing - Original Draft & \checkmark &  & \checkmark\\
    Writing - Review \& Editing & \checkmark & \checkmark  & \checkmark\\    
    Supervision & & \checkmark  & \checkmark\\\hline
\end{tabular}\par
}

\section{Sentence-level asymmetry effects}
\label{app:semsim_spe}

\begin{figure*}
    \centering
    \includegraphics[width=\textwidth]{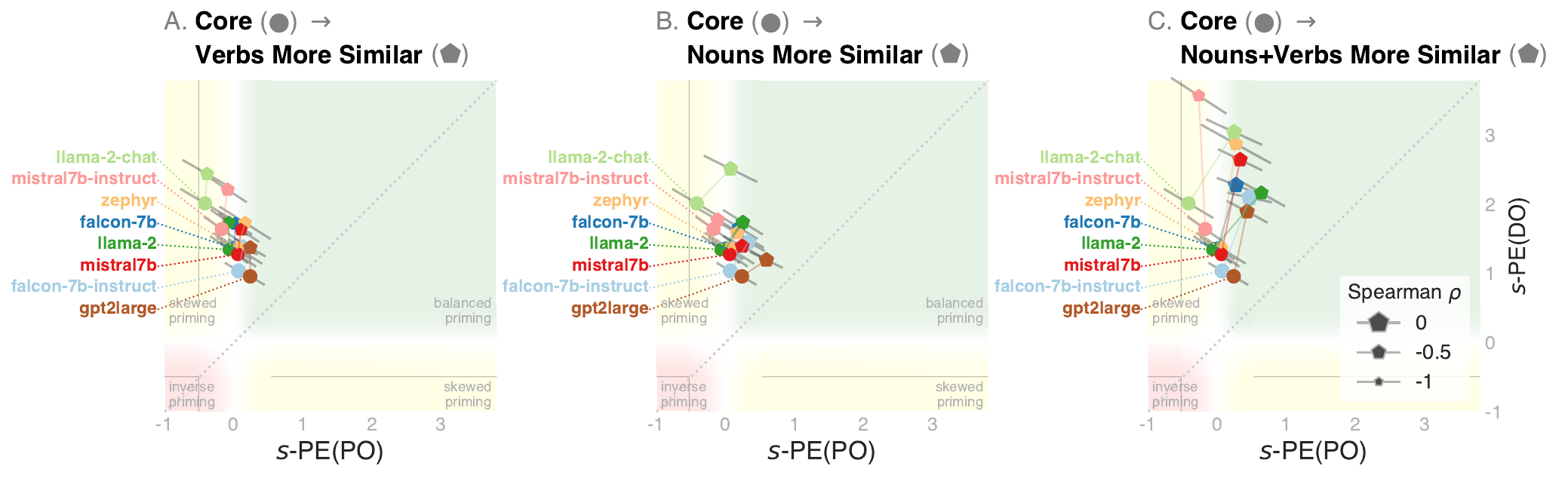}
    \caption{Sentence-level Priming Effects for conditions with an increased semantic similarity across nouns and verbs.}
    \label{fig:lexboost-semsim-pe-full}
\end{figure*}

The increased semantic similarity, in Figure~\ref{fig:lexboost-semsim-pe-full}, can be seen to primarily have an effect on boosting the PE score of the dominant structure (DO), while leaving the PE score of the opposite structure unaffected (PO).
Furthermore, it can be seen that the the effect of increasing the similarity of nouns and verbs is not \textit{linearly additive}: increasing similarity of both nouns and verbs has a far greater impact than the individual conditions combined.

\section{Token-level PE for Increased Semantic Similarity}\label{app:semsim_wpe}
\begin{figure*}
    \centering
    \includegraphics[width=\textwidth]{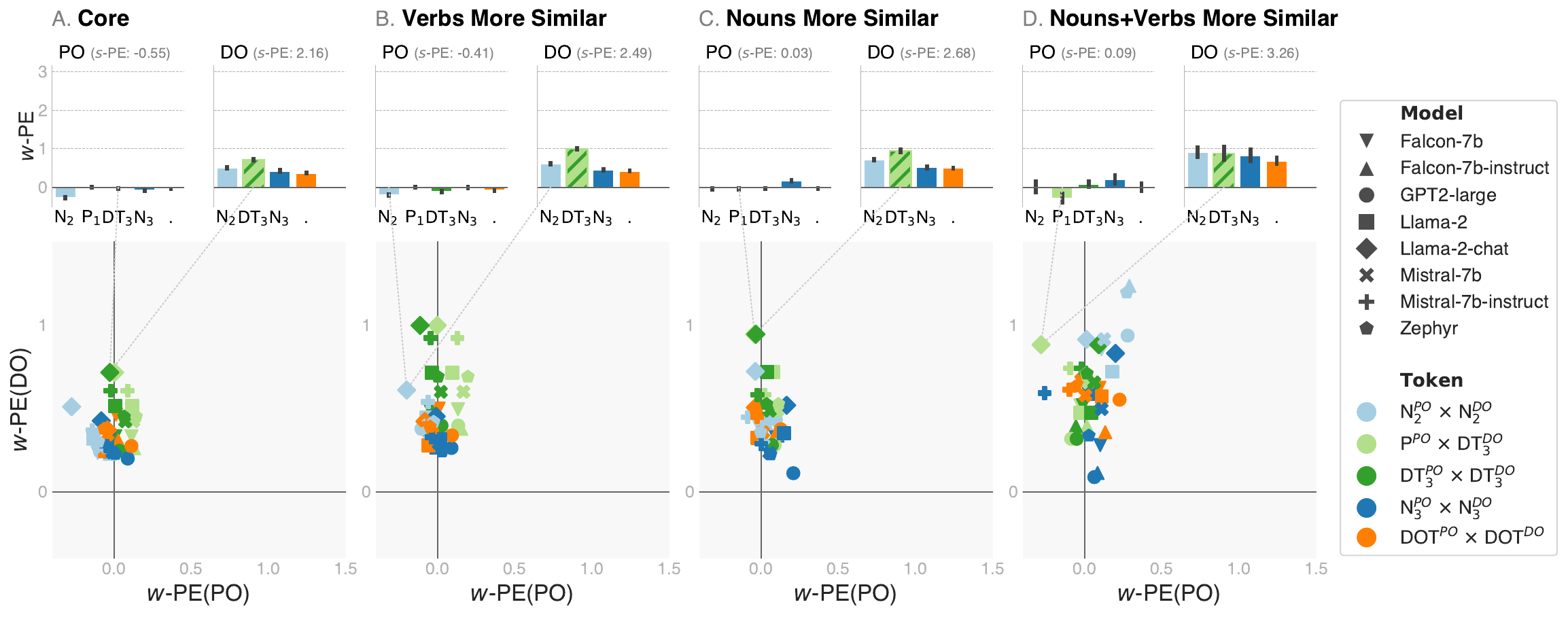}
    \caption{$w$-PE scores for increased semantic similarity between prime and target.}
    \label{fig:wpe-semsim}
\end{figure*}

The token-level PE scores for the 3 conditions with increased semantic similarity are shown in Figure~\ref{fig:wpe-semsim}.
\begin{figure*}
    \centering
    \includegraphics[width=0.9\textwidth]{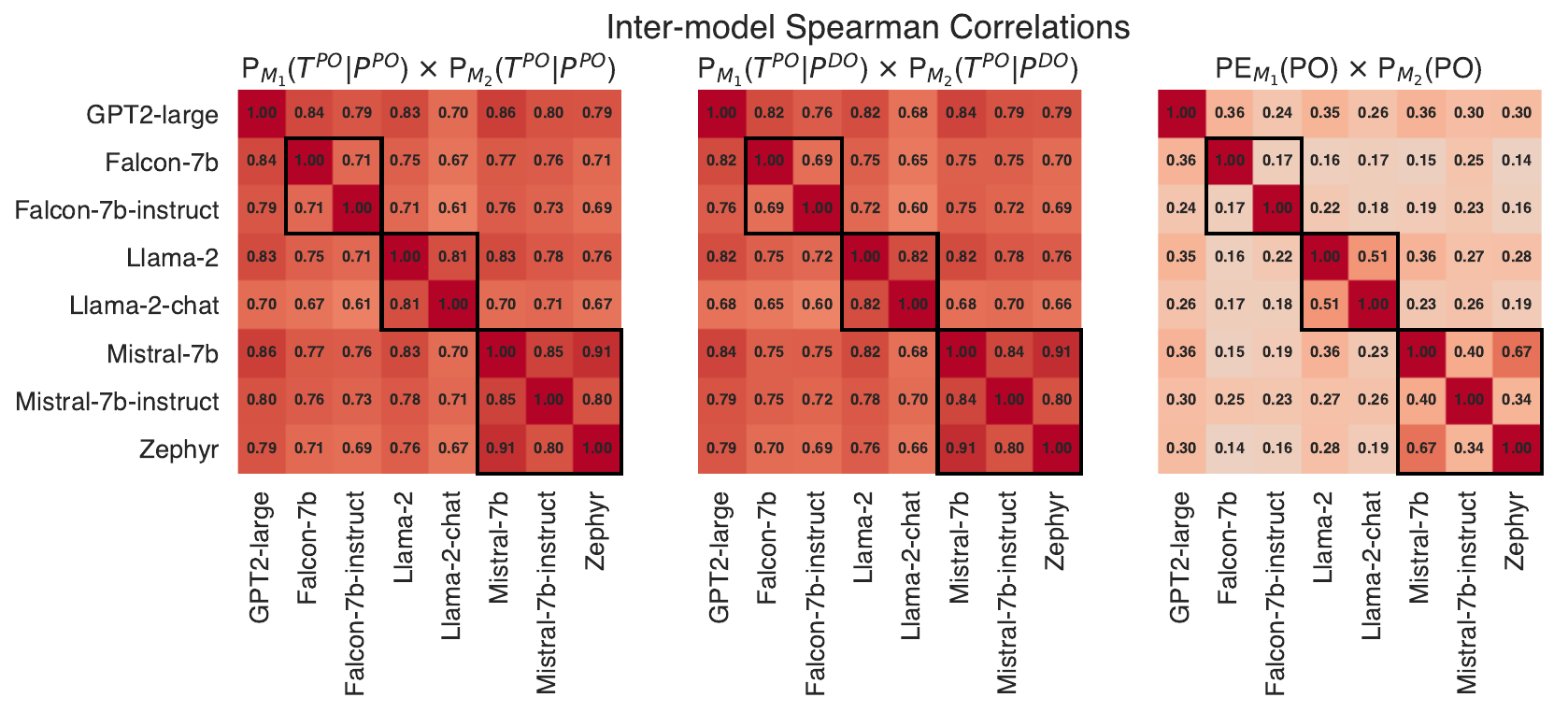}
    \caption{Correlations between LMs.}
    \label{fig:alignment_base_corr}
\end{figure*}

\section{Preference Order Correlation} \label{sec:structural_preference_correlations}
\begin{figure}[h!]
    \centering
    \includegraphics[width=\columnwidth]{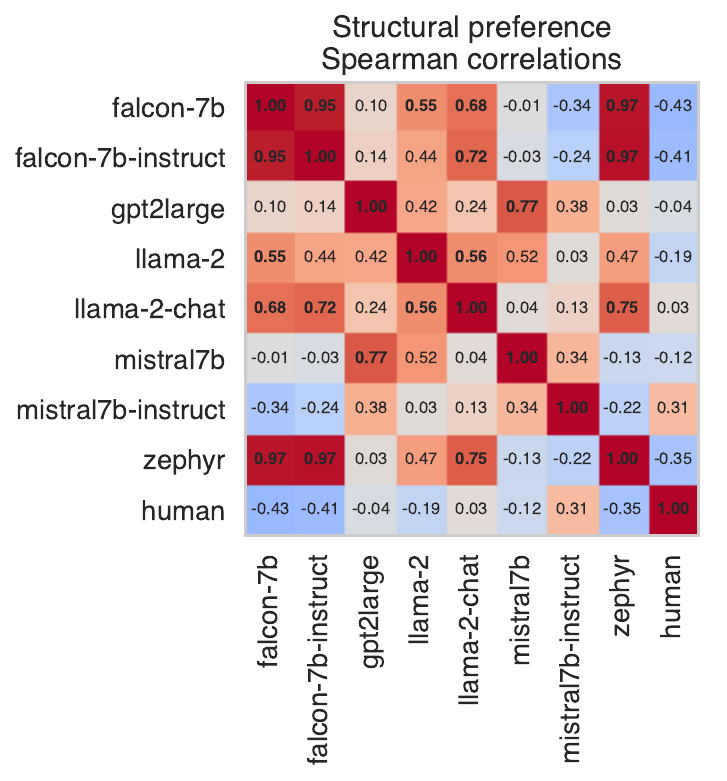}
    \caption{The Spearman correlation between LMs and human structural preference order.}
    \label{fig:structural_preference_correlations}
\end{figure}

The Spearman correlation between LMs and human structural preference order are shown in Figure~\ref{fig:structural_preference_correlations}.
Surprisingly, there exists a high degree of variation in preference order, both within models and across models and human preference.
Only \textit{Falcon-7b} and its instruction-tuned variant retain a high preference overlap; all the other aligned LMs diverge quite strongly from their base model.
None of the LMs have a significant correlation with respect to the reported human preference order, which is in contrast to \citet{hawkins-etal-2020-investigating}'s positive findings of strong correlations between model and human preferences.
We leave a more thorough investigation of these differences open for future work.



\section{Linear Mixed-effect Model Summary}\label{app:lmm_summary}
The LMM results including coefficients, standard error, $z$-score and $p$-values are shown in Table~\ref{tab:pe_po_results} and \ref{tab:pe_do_results}.

\begin{table*}
\centering
\small
\begin{tabular}{lllll}
\toprule
0 &             Model: &  MixedLM &  Dependent Variable: &  $s_\delta$-PE(\textsc{po}) \\
1 &  No. Observations: &    30000 &              Method: &         REML \\
2 &        No. Groups: &        8 &               Scale: &       1.8203 \\
3 &   Min. group size: &     3655 &      Log-Likelihood: &  -51620.4121 \\
4 &   Max. group size: &     3858 &           Converged: &           No \\
5 &   Mean group size: &   3750.0 &                      &              \\
\bottomrule
\end{tabular}\\[30pt]

\begin{tabular}{lllllll}
\toprule
{} &   Coef. & Std.Err. &        z &  P>|z| &  [0.025 &  0.975] \\
\midrule
Intercept           &  -0.072 &    0.111 &   -0.652 &  0.515 &  -0.289 &   0.145 \\
sim($n_1$)          &   0.041 &    0.010 &    4.260 &  0.000 &   0.022 &   0.060 \\
sim($n_2$)          &   0.097 &    0.009 &   10.381 &  0.000 &   0.079 &   0.115 \\
sim($n_3$)          &   0.109 &    0.009 &   11.652 &  0.000 &   0.091 &   0.127 \\
sim($v$)            &   0.010 &    0.009 &    1.092 &  0.275 &  -0.008 &   0.027 \\
sim($s$)            &  -0.000 &    0.017 &   -0.008 &  0.994 &  -0.034 &   0.034 \\
$N_1$ overlaps      &  -0.148 &    0.052 &   -2.850 &  0.004 &  -0.250 &  -0.046 \\
$N_2$ overlaps      &   0.696 &    0.056 &   12.510 &  0.000 &   0.587 &   0.805 \\
$N_3$ overlaps      &   0.473 &    0.050 &    9.405 &  0.000 &   0.374 &   0.571 \\
Det. overlaps       &   1.010 &    0.026 &   38.149 &  0.000 &   0.958 &   1.062 \\
Verb overlaps       &   1.491 &    0.046 &   32.578 &  0.000 &   1.401 &   1.581 \\
Prep. overlaps      &   1.025 &    0.027 &   38.574 &  0.000 &   0.973 &   1.077 \\
$-$P(prime$_{po}$)  &   0.390 &    0.023 &   16.788 &  0.000 &   0.345 &   0.436 \\
$-$P(prime$_{do}$)  &  -0.259 &    0.025 &  -10.504 &  0.000 &  -0.307 &  -0.210 \\
$-$P(target$_{po}$) &   0.044 &    0.023 &    1.885 &  0.059 &  -0.002 &   0.089 \\
$-$P(target$_{do}$) &   0.055 &    0.025 &    2.223 &  0.026 &   0.006 &   0.103 \\
PO-pref($v^p$)      &  -0.129 &    0.010 &  -13.377 &  0.000 &  -0.148 &  -0.110 \\
PO-pref($v^t$)      &  -0.029 &    0.010 &   -3.041 &  0.002 &  -0.048 &  -0.010 \\
Group Var           &   0.097 &    0.061 &          &        &         &         \\
\bottomrule
\end{tabular}
\caption{Raw LMM results for predicting $s_\delta$-PE(\textsc{po}).}\label{tab:pe_po_results}
\end{table*}

\begin{table*}
\centering
\small
\begin{tabular}{lllll}
\toprule
0 &             Model: &  MixedLM &  Dependent Variable: &   $s_\delta$-PE(\textsc{do}) \\
1 &  No. Observations: &    30000 &              Method: &         REML \\
2 &        No. Groups: &        8 &               Scale: &       2.2337 \\
3 &   Min. group size: &     3655 &      Log-Likelihood: &  -54688.2103 \\
4 &   Max. group size: &     3858 &           Converged: &           No \\
5 &   Mean group size: &   3750.0 &                      &              \\
\bottomrule
\end{tabular}
\\[30pt]

\begin{tabular}{lllllll}
\toprule
{} &   Coef. & Std.Err. &        z &  P>|z| &  [0.025 &  0.975] \\
\midrule
Intercept           &   1.339 &    0.111 &   12.047 &  0.000 &   1.121 &   1.557 \\
sim($n_1$)          &   0.071 &    0.011 &    6.701 &  0.000 &   0.050 &   0.092 \\
sim($n_2$)          &   0.021 &    0.010 &    2.018 &  0.044 &   0.001 &   0.041 \\
sim($n_3$)          &   0.122 &    0.010 &   11.826 &  0.000 &   0.102 &   0.142 \\
sim($v$)            &   0.171 &    0.010 &   17.197 &  0.000 &   0.151 &   0.190 \\
sim($s$)            &  -0.044 &    0.019 &   -2.331 &  0.020 &  -0.082 &  -0.007 \\
$N_1$ overlaps      &   0.456 &    0.057 &    7.966 &  0.000 &   0.344 &   0.568 \\
$N_2$ overlaps      &  -0.146 &    0.061 &   -2.384 &  0.017 &  -0.266 &  -0.026 \\
$N_3$ overlaps      &   0.717 &    0.055 &   12.947 &  0.000 &   0.608 &   0.826 \\
Det. overlaps       &   1.671 &    0.029 &   57.289 &  0.000 &   1.614 &   1.728 \\
Verb overlaps       &   1.535 &    0.050 &   30.433 &  0.000 &   1.436 &   1.634 \\
Prep. overlaps      &   0.244 &    0.029 &    8.349 &  0.000 &   0.187 &   0.302 \\
$-$P(prime$_{po}$)  &  -0.333 &    0.026 &  -12.989 &  0.000 &  -0.383 &  -0.283 \\
$-$P(prime$_{do}$)  &   0.416 &    0.027 &   15.332 &  0.000 &   0.363 &   0.469 \\
$-$P(target$_{po}$) &  -0.045 &    0.026 &   -1.759 &  0.079 &  -0.095 &   0.005 \\
$-$P(target$_{do}$) &   0.311 &    0.027 &   11.452 &  0.000 &   0.258 &   0.365 \\
PO-pref($v^p$)      &   0.231 &    0.011 &   21.761 &  0.000 &   0.210 &   0.252 \\
PO-pref($v^t$)      &   0.215 &    0.011 &   20.262 &  0.000 &   0.194 &   0.236 \\
Group Var           &   0.097 &    0.031 &          &        &         &         \\
\bottomrule
\end{tabular}
\caption{Raw LMM results for predicting $s_\delta$-PE(\textsc{do}).}\label{tab:pe_do_results}
\end{table*}

\section{Model Size and Alignment}
\label{sec:modelsizealignment}

Interestingly, although the sample is too small to make broad generalisations, in the models we test, we observe larger models exhibit more \textit{skewed} priming behaviour in the core, and higher susceptibility to lexical boosting than the smaller GPT2. 
We also observe no strong patterns to distinguish alignment tuning from base models, in fact, one surprising finding is the degree of difference in PE for a given prime target pair that a base and alignment model from the same base will have (See Figure~\ref{fig:alignment_base_corr}, which has the models grouped by base model).

\end{document}